\documentclass[letterpaper, 10pt, conference]{ieeeconf}  

\IEEEoverridecommandlockouts                              

\usepackage{cite}
\usepackage{amsmath,amssymb,amsfonts}
\usepackage{algorithmic}
\usepackage{algorithm}
\usepackage{graphicx}
\usepackage{textcomp}
\usepackage{xcolor}
\def\BibTeX{{\rm B\kern-.05em{\sc i\kern-.025em b}\kern-.08em
    T\kern-.1667em\lower.7ex\hbox{E}\kern-.125emX}}
\usepackage{booktabs}
\usepackage{multirow}
\usepackage[super]{nth}
\usepackage{tikz}
\newcommand*\circled[1]{\tikz[baseline=(char.base)]{
		\node[shape=circle,draw,inner sep=0.2pt] (char) {#1};}}
\newcommand*\circledB[1]{\tikz[baseline=(char.base)]{
            \node[shape=circle,fill,inner sep=0.2pt] (char) {\textcolor{white}{#1}};}}
\usetikzlibrary{tikzmark}
\tikzset{circledColor/.style={circle,draw,inner sep=0.1em,line width=0.04em}}
\usepackage{soul}
\usepackage{array}
\newcolumntype{?}{!{\vrule width 1.5pt}}

\usepackage{setspace}
\usepackage{cleveref}

\title{\LARGE \bf
NeuroNAS: Enhancing Efficiency of Neuromorphic In-Memory Computing for Intelligent Mobile Agents through Hardware-Aware Spiking Neural Architecture Search
}

\author{
Rachmad Vidya Wicaksana Putra and Muhammad Shafique
\thanks{Rachmad Vidya Wicaksana Putra is with eBrain Lab, New York University (NYU) Abu Dhabi, Abu Dhabi, United Arab Emirates.
{\tt\footnotesize Email: rachmad.putra@nyu.edu}}%
\thanks{Muhammad Shafique is the Director of eBrain Lab, New York University (NYU) Abu Dhabi, Abu Dhabi, United Arab Emirates.
{\tt\footnotesize Email: muhammad.shafique@nyu.edu}}%
}

\begin{document}

\maketitle
\thispagestyle{empty}
\pagestyle{empty}

\begin{abstract}
Intelligent mobile agents (e.g., UGVs and UAVs) typically demand low power/energy consumption when solving their machine learning (ML)-based tasks, since they are usually powered by portable batteries with limited capacity. 
A potential solution is employing neuromorphic computing with Spiking Neural Networks (SNNs), which leverages event-based computation to enable ultra-low power/energy ML algorithms. 
To maximize the performance efficiency of SNN inference, the In-Memory Computing (IMC)-based hardware accelerators with emerging device technologies (e.g., RRAM) can be employed. 
However, SNN models are typically developed without considering constraints from the application and the underlying IMC hardware, thereby hindering SNNs from reaching their full potential in performance and efficiency. 
To address this, we propose \textit{NeuroNAS}, a novel framework for developing energy-efficient neuromorphic IMC for intelligent mobile agents using hardware-aware spiking neural architecture search (NAS), i.e., by quickly finding an SNN architecture that offers high accuracy under the given constraints (e.g., memory, area, latency, and energy consumption). 
Its key steps include: optimizing SNN operations to enable efficient NAS, employing quantization to minimize the memory footprint, developing an SNN architecture that facilitates an effective learning, and devising a systematic hardware-aware search algorithm to meet the constraints.
Compared to the state-of-the-art techniques, NeuroNAS quickly finds SNN architectures (with 8bit weight precision) that maintain high accuracy by up to 6.6x search time speed-ups, while achieving up to 92\% area savings, 1.2x latency improvements, 84\% energy savings across different datasets (i.e., CIFAR-10, CIFAR-100, and TinyImageNet-200); while the state-of-the-art fail to meet all constraints at once. 
Therefore, NeuroNAS enables efficient design automation in developing neuromorphic IMC for energy-efficient mobile agents.
\end{abstract}

\section{Introduction}
\label{Sec_Intro}

\begin{figure}[t]
\centering
\includegraphics[width=\linewidth]{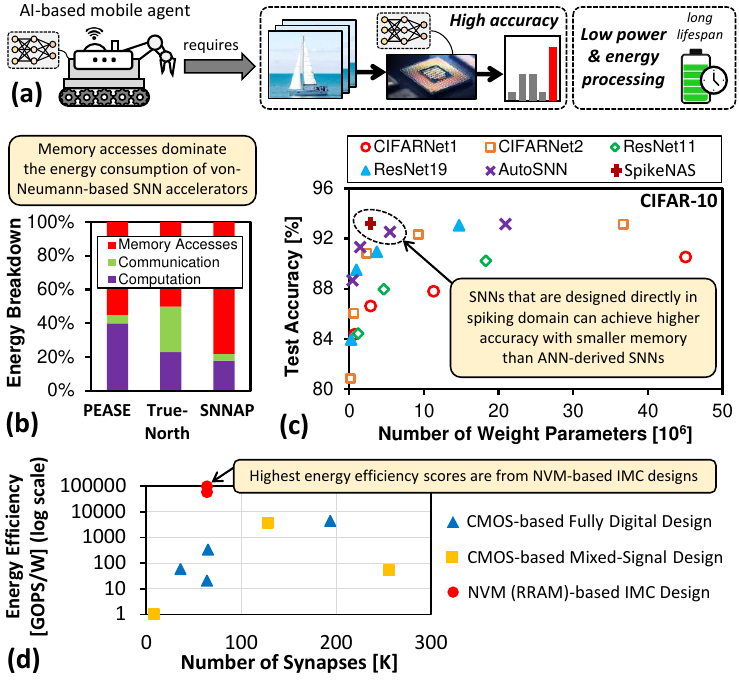}
\vspace{-0.5cm}
\caption{\textbf{(a)} Intelligent mobile agents are typically expected to provide high accuracy with low power/energy processing to solve ML-based tasks.
\textbf{(b)} Breakdown of SNN inference energy on von-Neumann-based neuromorphic accelerators: PEASE~\cite{Ref_Roy_PEASE_ISLPED17}, TrueNorth~\cite{Ref_Akopyan_TrueNorth_TCAD15}, and SNNAP~\cite{Ref_Sen_ApproxSNN_DATE17}; adapted from~\cite{Ref_Krithivasan_SpikeBundle_ISLPED19}. 
\textbf{(c)} Accuracy vs. memory of different SNNs on CIFAR-10 dataset: CIFARNet1~\cite{Ref_Wu_DirectTrainSNNs_AAAI19}, CIFARNet2~\cite{Ref_Fang_MemTConstantSNNs_ICCV21}, ResNet11~\cite{Ref_Lee_SpikeBackprop_FNINS20}, ResNet19~\cite{Ref_Zheng_LargerSNNs_AAAI21}, AutoSNN~\cite{Ref_Na_AutoSNN_ICML22}, and SpikeNAS~\cite{Ref_Putra_SpikeNAS_arXiv24}; data are obtained from studies in~\cite{Ref_Na_AutoSNN_ICML22}\cite{Ref_Putra_SpikeNAS_arXiv24}.
\textbf{(d)} Energy efficiency of different types of single-core SNN accelerators: CMOS-based fully digital design, CMOS-based mixed-signal design, and RRAM-based IMC design; based on data from~\cite{Ref_Basu_SNNicSurvey_CICC22}.}
\label{Fig_MotivationIMC}
\vspace{-0.5cm}
\end{figure}

Intelligent mobile agents (e.g., UGVs and UAVs) have shown potentials in helping humans in solving diverse, complex, and intensive machine learning (ML) tasks in more accurate and efficient way~\cite{Ref_Putra_EmbodiedNAI_ICARCV24}. 
Apart from high accuracy, these agents require low power/energy execution of ML algorithms as they are usually powered by limited portable batteries; see Fig.~\ref{Fig_MotivationIMC}(a). 
A potential solution is by employing neuromorphic computing with Spiking Neural Networks (SNNs) which can offer ultra-low power/energy ML algorithms through their very sparse event-based operations.

To maximize the efficiency of SNN inference, specialized accelerators have been employed~\cite{Ref_Basu_SNNicSurvey_CICC22}.
However, they suffer from high memory access energy which dominates the overall system energy (\textit{memory wall bottleneck}); see Fig.~\ref{Fig_MotivationIMC}(b).
To address this, the non-von Neumann In-Memory Computing (IMC) paradigm with non-volatile memory (NVM) technologies, such as resistive random access memory (RRAM), has been investigated~\cite{Ref_Ankit_Resparc_DAC17}\cite{Ref_Moitra_SpikeSim_TCAD23}.
Specifically, IMC platforms minimize data movements between the memory and compute engine parts, hence the memory access energy; see Fig.~\ref{Fig_MotivationIMC}(d).

However, simply executing the existing SNN models on an IMC platform may not fulfill the design requirements of mobile agents in terms of performance (e.g., accuracy and latency) and hardware constraints (e.g., memory, area, and energy consumption), as SNN models are usually developed without considering the characteristics of the IMC hardware. 
Moreover, most of SNN models are derived from conventional Artificial Neural Networks (ANNs), whose operations are different from SNNs, thus they may lose unique SNN features (e.g., temporal information) that leads to sub-optimal accuracy~\cite{Ref_Kim_SNASNet_ECCV22}.
This limitation is also observed in recent studies~\cite{Ref_Putra_SpikeNAS_arXiv24}\cite{Ref_Kim_SNASNet_ECCV22}, i.e., designing SNNs in the spiking domain can achieve higher accuracy than deriving SNNs from the ANN domain, as shown in Fig.~\ref{Fig_MotivationIMC}(c). 
Furthermore, to ensure the practicality of SNNs for real-world implementation of mobile agents, SNN developments have to meet the application requirements and constraints.
In fact, \textit{maximizing SNN benefits for mobile agent applications and IMC platforms requires SNN developments that can meet the expected performance, efficiency, and constraints}, which is a non-trivial task. 
Moreover, manually developing the suitable SNN is time consuming and laborious, hence requiring an alternative technique that can find a desired solution efficiently.

\textbf{Research Problem:} 
\textit{How to automatically and quickly develop an SNN architecture for neuromorphic-based IMC that achieves high accuracy and efficiency under the given application requirements and constraints?}
An efficient solution to this problem may enable automatic and quick developments of efficient neuromorphic IMC for intelligent mobile agents.  

\subsection{State-of-the-Art of NAS for SNNs and Their Limitations}
\label{Sec_Intro_SOTA}

Currently, state-of-the-art works focus on employing \textit{neural architecture search (NAS)} for finding SNN architectures that achieve high accuracy~\cite{Ref_Na_AutoSNN_ICML22, Ref_Kim_SNASNet_ECCV22, Ref_Putra_SpikeNAS_arXiv24, Ref_Liu_litESNN_arXiv24, Ref_Yan_NAS4SNN_NeuNet24, Ref_Pan_EvoNAS4SNN_arxiv23, Ref_Che_AutoSpikeformer_arXiv23, Ref_Wang_AutoST_ICASSP24}. 
However, \textit{these NAS methods do not consider multiple design requirements and constraints posed by the targeted application and the IMC hardware (i.e., memory, area, latency, and energy consumption)}, thereby limiting the applicability of neuromorphic IMC for real-world mobile agents. 
To illustrate the limitations of state-of-the-art and the optimization potentials, a case study is performed and discussed in Sec.~\ref{Sec_Intro_Challenges}. 

\subsection{Case Study and Associated Research Challenges}
\label{Sec_Intro_Challenges}

\begin{figure}[h]
\vspace{-0.5cm}
\centering
\includegraphics[width=\linewidth]{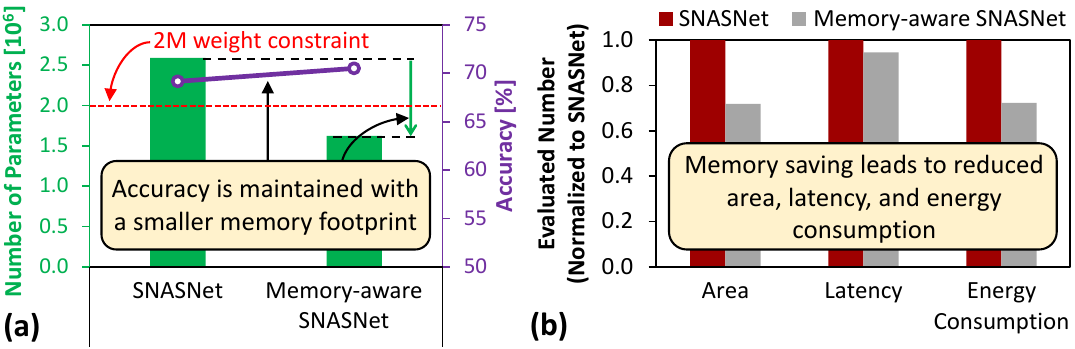}
\vspace{-0.6cm}
\caption{Results from the case study considering the CIFAR-100 dataset: \textbf{(a)} number of weight parameters and accuracy; and \textbf{(b)} area, latency, and energy consumption when running SNNs on an RRAM-based IMC hardware~\cite{Ref_Moitra_SpikeSim_TCAD23}.}
\label{Fig_ObserveConstraints}
\vspace{-0.3cm}
\end{figure}

We study the impact of applying constraints in the NAS process. 
To do this, we reproduce the state-of-the-art NAS for SNNs (i.e., SNASNet~\cite{Ref_Kim_SNASNet_ECCV22}), and then apply a memory constraint of 2x10$^6$ (2M) weight parameters on it to build the memory-aware SNASNet, while considering the CIFAR-100 dataset.
For the hardware, we consider an RRAM-based IMC platform from~\cite{Ref_Moitra_SpikeSim_TCAD23}.
Note, details of experimental setup are discussed in Sec.~\ref{Sec_Eval}.
The experimental results are shown in Fig.~\ref{Fig_ObserveConstraints}, from which we make the following key observations.
\begin{enumerate}
    \item A smaller SNN may achieve comparable accuracy to the larger one, and hence showing a possibility to obtain high accuracy with low memory.
    \item Memory saving leads to the reduction of area, latency, and energy consumption of SNN processing. 
\end{enumerate}

From these observations, we identify \textbf{research challenges} to be addressed for solving the targeted problem, as follows.
\begin{itemize}
    \item The solution should consider multiple design requirements and constraints (i.e., memory, area, latency, and energy consumption) in its search algorithm to ensure the deployability of SNNs on the IMC platform for mobile agents. 
    \item The solution should quickly perform its searching process to minimize the searching time. 
\end{itemize}

\subsection{Our Novel Contributions}
\label{Sec_Intro_NovelContrib}

\begin{figure}[h]
\vspace{-0.2cm}
\centering
\includegraphics[width=\linewidth]{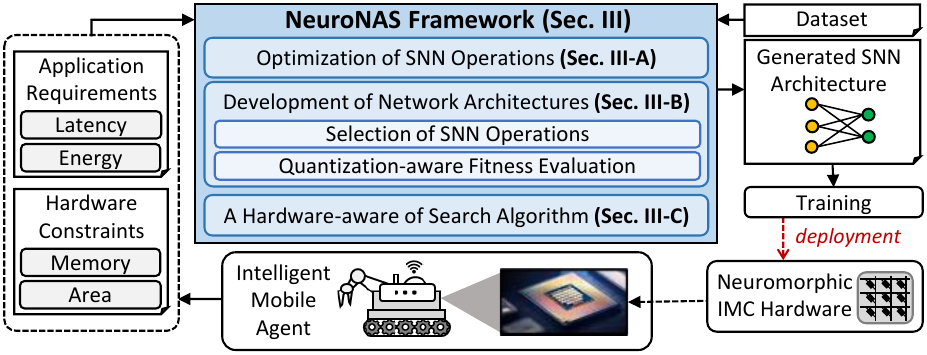}
\vspace{-0.5cm}
\caption{Overview of our novel contributions in the NeuroNAS framework.}
\label{Fig_NeuroNAS_Novelty}
\end{figure}

To address the targeted problem and research challenges, we propose \textit{\textbf{NeuroNAS}, a novel framework for quickly developing the SNN architecture that can achieve high accuracy, while meeting the application requirements and IMC constraints (i.e., memory, area, latency, and energy consumption) through hardware-aware spiking NAS}. 
It is also the first work that develops a hardware-aware spiking NAS for enhancing efficient neuromorphic IMC design targeting intelligent mobile agents.   
NeuroNAS employs the following key steps (see an overview in Fig.~\ref{Fig_NeuroNAS_Novelty} and details in Fig.~\ref{Fig_NeuroNAS_Framework}). 
\begin{enumerate}
    \item \textbf{Optimization of SNN Operations (Sec.~\ref{Sec_NeuroNAS_SNNops}):} 
    It investigates then selects SNN operations that have positive impact on the accuracy, while considering their memory costs. 
    These operations will be used for building SNNs. 
    \item \textbf{Development of Network Architectures (Sec.~\ref{Sec_NeuroNAS_SNNarch}):} 
    It aims to build an SNN architecture that facilitates effective learning and efficient deployments on IMC platform. 
    It leverages the selected operations, and the quantization-aware fitness evaluation policy to improve the quality of fitness function and optimize the network size.
    \item \textbf{A Hardware-aware Search Algorithm (Sec.~\ref{Sec_NeuroNAS_SearchAlg}):} 
    It aims to find the suitable SNN architecture that achieves high accuracy, while meeting the given requirements and constraints on memory, area, latency, and energy. 
\end{enumerate}  

\textbf{Key Results:} 
We evaluate NeuroNAS framework through a PyTorch implementation, and run it on Nvidia RTX A6000 Multi-GPU machines. 
As a result, NeuroNAS quickly finds SNNs (with 8bit weight precision) that maintain high accuracy by up to 6.6x search time speed-ups, while achieving up to 92\% area savings, 1.2x latency improvements, 84\% energy savings across CIFAR-10, CIFAR-100, and TinyImageNet-200 compared to the state-of-the-art. 
Meanwhile, the state-of-the-art fail to meet all constraints at once.

\section{Preliminaries}
\label{Sec_Prelim}
 
\subsection{Spiking Neural Networks (SNNs)}
\label{Sec_Prelim_SNNs}

SNNs draw inspiration from the biological neurons, i.e., utilizing spikes for conveying information and processing data~\cite{Ref_Putra_SparkXD_DAC21, Ref_Putra_SpikeDyn_DAC21, Ref_Putra_ReSpawn_ICCAD21}. 
Therefore, operations in SNNs depend on the neuronal dynamics of the spiking neuron model. 
Typically, a leaky integrate-and-fire (LIF) neuron model is used since it can provide diverse spike patterns with a low-complexity. 

\begin{figure}[t]
\centering
\includegraphics[width=\linewidth]{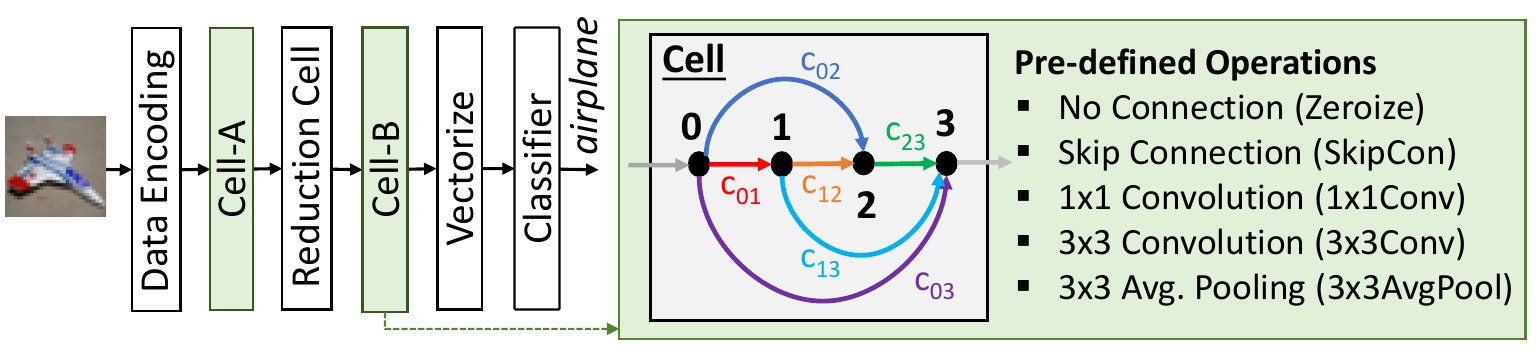}
\vspace{-0.7cm}
\caption{SNN macro-architecture with 2 neural cells, and each cell is a DAG whose edge denotes a specific pre-defined operation; adapted from~\cite{Ref_Kim_SNASNet_ECCV22}.}
\label{Fig_MacroArch}
\vspace{-0.2cm}
\end{figure}

\begin{figure}[t]
\centering
\includegraphics[width=\linewidth]{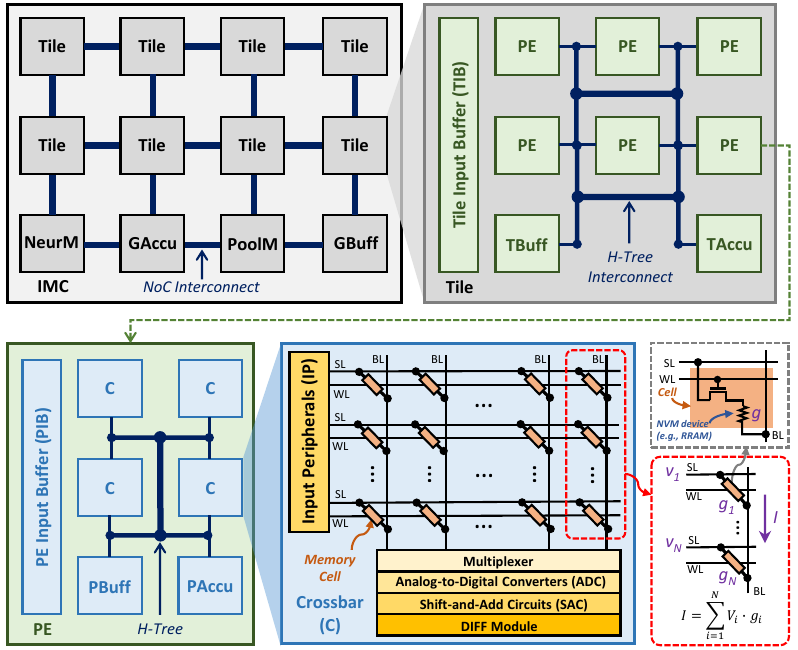}
\vspace{-0.6cm}
\caption{The SNN IMC hardware architecture that is considered in this work. It is based on the SpikeFlow architecture~\cite{Ref_Moitra_SpikeSim_TCAD23}.}
\label{Fig_CIMArch}
\vspace{-0.4cm}
\end{figure}

\begin{figure}[t]
\centering
\includegraphics[width=\linewidth]{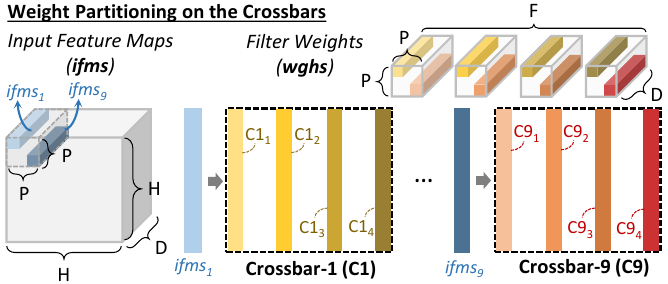}
\vspace{-0.7cm}
\caption{SNN partitioning for mapping input feature maps (\textit{ifms}) and weights (\textit{wghs}) on the IMC crossbars; adapted from studies in~\cite{Ref_Moitra_SpikeSim_TCAD23}.
The dimension of \textit{ifms} is denoted as $H$$\times$$H$$\times$$D$, while \textit{wghs} is denoted as $P$$\times$$P$$\times$$D$$\times$$F$.}
\label{Fig_MappingOnCIM}
\vspace{-0.4cm}
\end{figure}

\begin{figure*}[t]
\centering
\includegraphics[width=\linewidth]{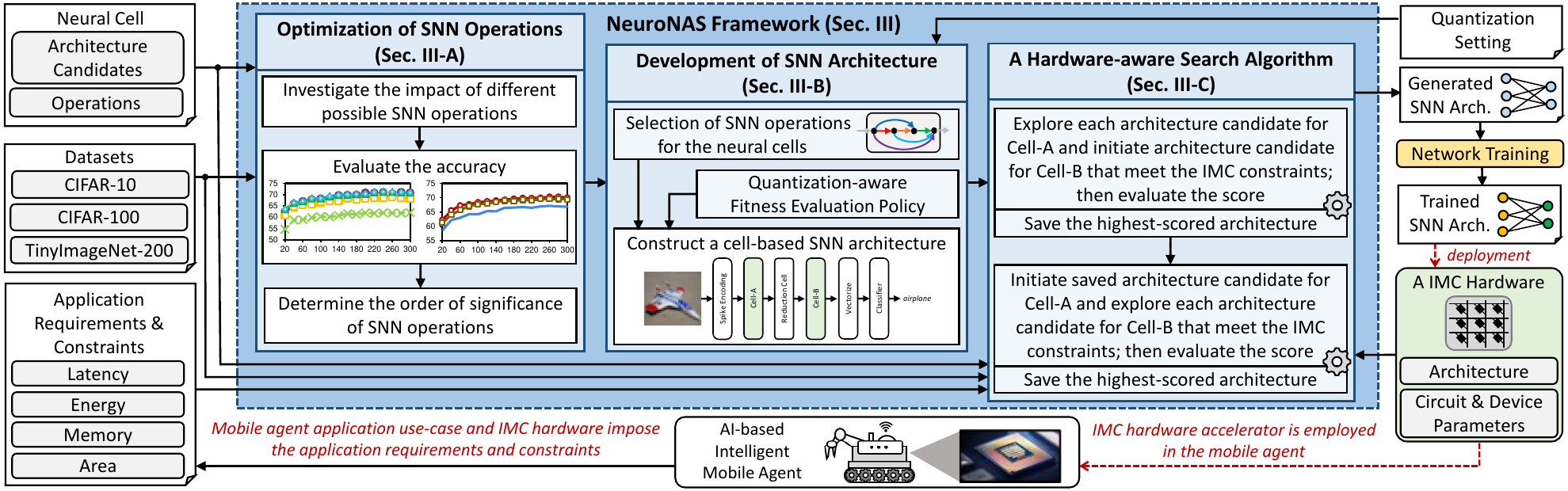}
\vspace{-0.5cm}
\caption{Overview of our NeuroNAS framework, and its key steps for generating an SNN architecture that can achieve high accuracy while meeting the given memory, area, latency, and energy constraints from the mobile agent with neuromorphic IMC hardware; novel contributions are highlighted in blue.}
\label{Fig_NeuroNAS_Framework}
\vspace{-0.2cm}
\end{figure*}

\subsection{NAS for SNNs}
\label{Sec_Prelim_NAS}

\textbf{Overview:} 
State-of-the-art works still focus on the NAS development for finding SNN architectures that achieve high accuracy. 
They can be categorized into two approaches.
First is \textit{``NAS with Training''}, which trains a super-network that contains all possible architecture candidates, and then search a subset of architecture that meets the fitness criteria, as demonstrated in~\cite{Ref_Na_AutoSNN_ICML22}\cite{Ref_Yan_NAS4SNN_NeuNet24}.
However, it incurs huge searching time, memory, and energy consumption. 
Second is \textit{``NAS without Training''}, which directly evaluates the fitness scores of architecture candidates, and then only trains the one with the highest score~\cite{Ref_Kim_SNASNet_ECCV22, Ref_Putra_SpikeNAS_arXiv24, Ref_Liu_litESNN_arXiv24}\cite{Ref_Pan_EvoNAS4SNN_arxiv23, Ref_Che_AutoSpikeformer_arXiv23, Ref_Wang_AutoST_ICASSP24}. 
However, \textit{all of these state-of-the-art methods have not considered multiple design requirements and constraints posed by the target application and the CIM platform}, hence limiting their applicability for real-world mobile agent applications. 

\smallskip
\textbf{NAS without Training:}
In this work, we consider \textit{``NAS without training''} approach~\cite{Ref_Kim_SNASNet_ECCV22, Ref_Putra_SpikeNAS_arXiv24, Ref_Liu_litESNN_arXiv24}\cite{Ref_Pan_EvoNAS4SNN_arxiv23, Ref_Che_AutoSpikeformer_arXiv23, Ref_Wang_AutoST_ICASSP24} due to its efficiency in finding an appropriate network without costly training, as compared to \textit{``NAS with Training''} approach~\cite{Ref_Na_AutoSNN_ICML22}\cite{Ref_Yan_NAS4SNN_NeuNet24}.
This concept was initially proposed for ANNs~\cite{Ref_Mellor_NASnoTraining_ICML21}, and then adopted in spiking domain.
For instance, SNASNet~\cite{Ref_Kim_SNASNet_ECCV22} leverages the studies in~\cite{Ref_Mellor_NASnoTraining_ICML21} for evaluating the SNN based on its representation capabilities, which indicate the potentials in learning effectively. 
It employs the following key steps. 
\begin{enumerate}
    \item It feeds mini-batch samples ($S$) to network, then records the LIF neurons' activities. 
    If a neuron generates a spike, it is mapped to 1, and otherwise 0. 
    In each layer, this mapping is represented as binary vector ($b$).
    \item The Hamming distance $H(b_i, b_j)$ between samples $i$ and $j$ is calculated to construct a matrix ($\textbf{M}_H$); see Eq.~\ref{Eq_Kmatrix}. 
    $N$ is the number of neurons in the investigated layer, while $\beta$ is the normalization factor to address high sparsity issue. 
    \item The fitness score ($F$) of each investigated architecture candidate is calculated using $F$=$\log \left( \det \; \lvert \sum_l \textbf{M}_H^l  \rvert \right)$, and the one with highest-score is selected for training.
    \smallskip
\end{enumerate}
\begin{equation}
    \small
    \textbf{M}_H = 
    \begin{pmatrix}
    N - \beta H(b_1,b_1) & \cdots & N - \beta H(b_1,b_S) \\
    \vdots   & \ddots & \vdots  \\
    N - \beta H(b_S,b_1) & \cdots & N - \beta H(b_S,b_S) 
    \end{pmatrix}
    \label{Eq_Kmatrix}
\end{equation}   

\noindent In this work, \textit{we propose a similar approach but enhanced with quantization-aware mechanism in our NeuroNAS framework to adapt to the IMC hardware configuration}, which will be discussed further in Sec.~\ref{Sec_NeuroNAS_SNNarch_MacroArch}. 

\smallskip
\textbf{Neural Cell-based Strategy for NAS:}
It aims to provide a unified benchmark for NAS algorithms~\cite{Ref_Dong_NASbench201_ICLR20}. 
A \textit{neural cell} (or simply \textit{cell}) is a directed acyclic graph (DAG), that is originally designed with 4 nodes and 5 pre-defined operations, 
and its 2 nodes are connected to each other via a specific pre-defined operation~\cite{Ref_Kim_SNASNet_ECCV22}\cite{Ref_Dong_NASbench201_ICLR20}, as shown in Fig.~\ref{Fig_MacroArch}. 
Its idea is to search for the cell architecture and operations. 
In this work, \textit{we employ a similar neural cell-based strategy in our NeuroNAS framework but optimized for enabling a quick search and high accuracy}, which will be discussed in Sec.~\ref{Sec_NeuroNAS_SNNops} until Sec.~\ref{Sec_NeuroNAS_SNNarch_NeuralCell}. 

\subsection{IMC Hardware Architecture and SNN Mapping Strategy}
\label{Sec_Prelim_CIM}

\textbf{Architecture: }
We employ the state-of-the-art IMC hardware for SNNs (i.e., SpikeFlow~\cite{Ref_Moitra_SpikeSim_TCAD23}), whose architecture is shown in Fig.~\ref{Fig_CIMArch}. 
It comprises multiple compute tiles (Tiles), a neuron module (NeurM), a global accumulator (GAccu), a pooling module (PoolM), and a global buffer (GBuff).
These modules are connected to each other using network-on-chip (NoC). 
Each tile comprises a tile input buffer (TIB), processing elements (PEs), a tile buffer (TBuff), and a tile accumulator (TAccu). 
Each PE comprises a PE input buffer (PIB), a number of crossbars (C), PE buffer (PBuff), and PE accumulator (PAccu).
These modules are connected to each other using H-Tree connection.
Meanwhile, each crossbar comprises input peripherals (IP), an array of NVM devices, multiplexers, analog-to-digital converter (ADC), shift-and-add circuit (SAC), and DIFF module to perform signed multiply-and-accumulate (MAC) operations.

\smallskip
\textbf{SNN Mapping:}
We employ the state-of-the-art mapping strategy on the IMC hardware from studies in~\cite{Ref_Moitra_SpikeSim_TCAD23}; see Fig.~\ref{Fig_MappingOnCIM}.  
Its idea is to distribute weights across the crossbars, while exploiting the weight- and feature map-reuse. 
If we consider dimensions of $H$$\times$$H$$\times$$D$ for input feature maps (\textit{ifms}), $P$$\times$$P$$\times$$D$$\times$$F$ for weights (\textit{wghs}), and $X$$\times$$X$ for crossbar, then the mapping strategy can be described as follows. 
\begin{itemize}
    \item Weight kernel elements in $P$$\times$$P$ are mapped on different crossbars. For instance, a $3$$\times$$3$ weight kernel require 9 crossbars; see C1-C9.
    \item Elements from $1$$\times$$1$ weight kernel along the channel $D$ are mapped along a column of the crossbars (i.e., $1$$\times$$1$$\times$$D$); see C1$_1$ - C1$_9$.
    \item Different weight filters are mapped on different columns of the same crossbar. For instance, 4 weight filters will occupy 4 columns in the same crossbar; see C1$_1$ - C1$_4$.
    \item A similar data partitioning strategy is applied to \textit{ifms} over the $P$$\times$$P$$\times$$D$ block. 
\end{itemize}
Furthermore, this mapping strategy considers a design choice that a network layer can be mapped on multiple tiles, but multiple network layers cannot be mapped in one tile~\cite{Ref_Moitra_SpikeSim_TCAD23}.

\section{The NeuroNAS Framework}
\label{Sec_NeuroNAS}

We propose NeuroNAS framework, whose key steps are discussed below (see overview in Fig.~\ref{Fig_NeuroNAS_Framework}).

\vspace{-0.2cm}
\subsection{Optimization of SNN Operations}
\label{Sec_NeuroNAS_SNNops}
\vspace{-0.1cm}

\begin{figure}[t]
\centering
\includegraphics[width=\linewidth]{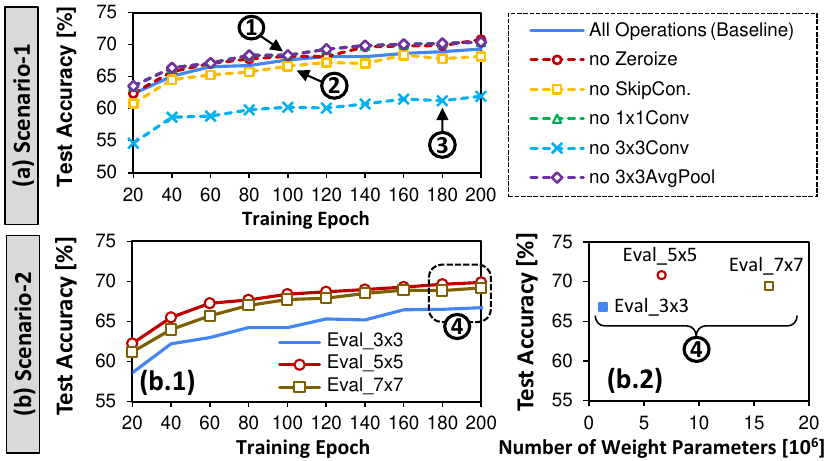}
\vspace{-0.4cm}
\caption{\textbf{(a)} Results of eliminating different operations from the search space of the cell architecture.
\textbf{(b)} Results of investigating the impact of 5x5Conv and 7x7Conv. Eval\_3x3, Eval\_5x5, and Eval\_7x7 refer to then accuracy when evaluating the impact of 3x3Conv, 5x5Conv, and 7x7Conv, respectively.}
\label{Fig_NeuroNAS_ObserveOps}
\vspace{-0.4cm}
\end{figure}

To develop an efficient SNN architecture that offers high accuracy, we have to understand the significance of its operations, then select the ones that lead to high accuracy with small memory cost. 
To do this, we perform an experimental case study with following scenarios.
\begin{itemize}
    \item \underline{Scenario-1:} 
    We remove an operation from the pre-defined ones (i.e., Zeroize, SkipCon, 1x1Conv, 3x3Conv, and 3x3AvgPool one-by-one), then perform NAS to evaluate the impact of their removal.
    \item \underline{Scenario-2:} 
    We investigate the impact of other possible operations with different kernel sizes (i.e., 5x5Conv and 7x7Conv) on accuracy and memory cost, to explore their potentials as alternative solutions. 
   Here, we replace the operation that has the role of extracting features from pre-define ones (i.e., 3x3Conv) with the investigated operation (either 5x5Conv or 7x7Conv).
\end{itemize}
All these scenarios consider a cell-based SNN illustrated in Fig.~\ref{Fig_MacroArch} and CIFAR-100.
Experimental results are shown in Fig.~\ref{Fig_NeuroNAS_ObserveOps}, from which we make the following key observations.

\begin{itemize}
    \item[\circled{1}] 
    Eliminating Zeroize, 1x1Conv, and 3x3AvgPool from the search space does not significantly degrade accuracy compared to the baseline. 
    Hence, these operations can be removed from the search space. 
    If model compression is needed for reducing the memory size, we can keep 3x3AvgPool in the search space. 
    \item[\circled{2}] 
    Eliminating SkipCon from the search space slightly reduces accuracy, since SkipCon is useful for providing feature maps from previous layer to preserve important information. 
    Similar results are also observed in~\cite{Ref_Putra_SpikeNAS_arXiv24}.
    Hence, SkipCon should be kept in the search space. 
    \item[\circled{3}] 
    Eliminating 3x3Conv from the search space significantly degrades accuracy, as the main role of 3x3Conv is to extract important features from input samples.  
    Hence, 3x3Conv should be kept in the search space to maintain the learning quality. 
    \item[\circled{4}] 
    Employing 5x5Conv and 7x7Conv can improve accuracy, since larger kernel sizes are capable of extracting more unique features from input samples.  
    However, they incur higher memory overheads, which lead to higher latency and energy consumption. 
\end{itemize}

These observations show the order of significance of the operations for achieving high accuracy with low memory cost, from the highest to the lowest: (1) 3x3Conv, (2) SkipCon, (3) 3x3AvgPool, then (4) 5x5Conv, 7x7Conv, Zeroize, and 1x1Conv.  
This insight is then leveraged in Sec.~\ref{Sec_NeuroNAS_SNNarch_NeuralCell}.

\subsection{Development of SNN Architecture}
\label{Sec_NeuroNAS_SNNarch}

This step aims to build an SNN architecture that facilitates effective learning.
It considers two parts: developments of (1) the neural cell architecture, and (2) the SNN architecture.

\subsubsection{\textbf{Neural Cell Architecture}}
\label{Sec_NeuroNAS_SNNarch_NeuralCell}

Based on our analysis in Sec.~\ref{Sec_NeuroNAS_SNNops}, we can optimize the search space of cell-based NAS strategy for maintaining/improving the accuracy while keeping the memory small. 
To do this, we select a few operations that have positive impact on accuracy with low memory cost to be considered in the search space. 
Specifically, \textit{we consider SkipCon, 3x3Conv, and 3x3AvgPool in the search space, while keeping 4 nodes of DAG in the neural cell}.

\subsubsection{\textbf{SNN Architecture}}
\label{Sec_NeuroNAS_SNNarch_MacroArch}

\begin{figure}[t]
\centering
\includegraphics[width=\linewidth]{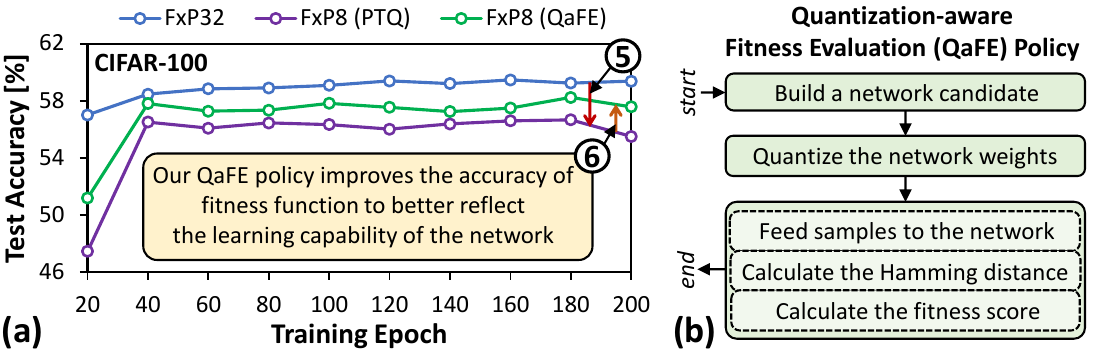}
\vspace{-0.5cm}
\caption{\textbf{(a)} Results of an experimental case study considering the CIFAR-100 dataset for networks with different fixed-point precision levels: 32-bit weights (FxP32), 8-bit weights with PTQ (FxP8-PTQ), and 8-bit with QaFE (FxP32-QaFE). \textbf{(b)} Key steps in our novel QaFE policy.}
\label{Fig_QAFE}
\end{figure}

The IMC platforms typically store SNN weights in NVM devices (e.g., RRAM) using a fixed-point format due to its efficient implementation~\cite{Ref_Moitra_SpikeSim_TCAD23}. 
Consequently, how each weight value stored in the IMC hardware depends on the NVM device precision.
For instance, a multi-bit weight value may be stored in several single-level-cell (SLC) devices, or in a single multi-level-cell (MLC) device. 
This indicates that SNNs should be developed and optimized considering the IMC parameters (e.g., NVM device precision) to ensure the deployability of SNNs. 
However, simply performing post-training quantization (PTQ) with fixed-point format may significantly degrade the accuracy; see \circled{5} in Fig.~\ref{Fig_QAFE}(a).
To address this, we propose a \textit{Quantization-aware Fitness Evaluation (QaFE) policy for quantizing the network candidates during NAS, thereby compressing the network size while improving the fitness function}.   
Our QaFE policy employs the following key steps; see Fig.~\ref{Fig_QAFE}(b).
\begin{itemize}
    \item An architecture candidate is built based on the given cell configuration (i.e., for topology and optimized operations).
    \item Weight quantization is then performed on the network based on the precision level and the rounding scheme.
    \item Then, mini-batch samples $S$ are fed to the network to calculate the Hamming distances between different samples, build a matrix $\textbf{M}_H$, and calculate the fitness score $F$.
\end{itemize}
This way, \textit{the fitness function is enhanced to better reflect the representation capability of the quantized network than the fitness score from the original network}; see \circled{6} in Fig.~\ref{Fig_QAFE}(a).

\subsection{A Hardware-aware Search Algorithm}
\label{Sec_NeuroNAS_SearchAlg}
\vspace{-0.1cm}

\textit{To systematically incorporate the proposed optimizations and perform the QaFE policy in the NAS process while meeting multiple design requirements and constraints, we propose a novel hardware-aware search algorithm}. 
Its pseudo-code is shown in Alg.~\ref{Alg_NeuroNAS}, whose key steps are discussed below.  
%
\begin{algorithm}[t]
  \footnotesize
  \caption{Our Hardware-aware Search Algorithm}
  \label{Alg_NeuroNAS}
  \begin{algorithmic}[1]
    \renewcommand{\algorithmicrequire}{\textbf{INPUT:}}
    \renewcommand{\algorithmicensure}{\textbf{OUTPUT:}}
	\REQUIRE 
        \textbf{(1)} Number of neural cell operations ($P$): SkipCon=0, 3x3Conv=1, 3x3AvgPool=2; 
        \textbf{(2)} Application requirements \& constraints: memory ($const_{mem}$), area ($const_{mem}$), latency ($const_{lat}$), energy consumption ($const_{eng}$);\\
	\ENSURE Generated an SNN architecture ($net\_arch$); \\
        \smallskip
	\textbf{BEGIN} \\
	  \textbf{Initialization}: \\
	  \STATE $score_h$=-1000; \\ 
        \smallskip
	  \textbf{Process}: \\
        \FOR{($i$=0; $i$$<$$2$; $i$++)}
            \FOR{each combination from ($\forall$ $c_{01}$ $\in$ $\{0, ..., P$-$1\}$), 
             ($\forall$ $c_{02}$ $\in$ $\{0, ..., P$-$1\}$), 
             ($\forall$ $c_{03}$ $\in$ $\{0, ..., P$-$1\}$), 
             ($\forall$ $c_{12}$ $\in$ $\{0, ..., P$-$1\}$),
             ($\forall$ $c_{13}$ $\in$ $\{0, ..., P$-$1\}$),
             ($\forall$ $c_{23}$ $\in$ $\{0, ..., P$-$1\}$)}
              \IF{($i$==0)}
                \STATE $cell_A$ = cell$(c_{01}, c_{02}, c_{03}, c_{12}, c_{13}, c_{23})$;
                \STATE $cell_B$ = $cell_A$;
                \STATE $arch$ = net$(cell_A, cell_B)$;
                \STATE $arch_q =$ quant$(arch)$; 
                \STATE $cost_{mem}$ = calc\_memory($arch_q$); 
                \IF{($cost_{mem} \leq const_{mem}$)}
                    \STATE $score =$ fitness\_eval$(arch_q)$; // evaluate the fitness score 
                    \IF{($score > score_h$)}
                        \STATE $score_h = score$;\\
                        \STATE $saved\_cell_A = cell_A$;\\
                        \STATE $saved\_cell_B = cell_B$;\\
                        \STATE $net\_arch = arch_q$;\\
                    \ENDIF\\
                \ENDIF\\
              \ENDIF\\
              \IF{($i$==1)}
                \STATE $cell_B$ = cell$(c_{01}, c_{02}, c_{03}, c_{12}, c_{13}, c_{23})$;
                \STATE $arch$ = net$(saved\_cell_A, cell_B)$;
                \STATE $arch_q =$ quant$(arch)$; 
                \STATE $cost_{mem}$ = calc\_memory($arch_q$); 
                \STATE $cost_{area}$= calc\_area($arch_q$); 
                \STATE $cost_{lat}$= calc\_latency($arch_q$);
                \STATE $cost_{eng}$= calc\_energy($arch_q$); 
                \IF{($cost_{mem} \leq const_{mem}$) and  
                    ($cost_{area} \leq const_{area}$) and 
                    ($cost_{lat} \leq const_{lat}$) and 
                    ($cost_{eng} \leq const_{eng}$)}
                        \STATE $score =$ fitness\_eval$(arch_q)$; // evaluate the fitness score
                        \IF{($score > score_h$)}
                            \STATE $score_h = score$;\\
                            \STATE $saved\_cell_B = cell_B$;\\
                            \STATE $net\_arch = arch_q$;\\
                        \ENDIF\\
                \ENDIF\\
              \ENDIF\\
            \ENDFOR\\
        \ENDFOR\\
	\RETURN $net\_arch$; \\
	\textbf{END}
	\end{algorithmic} 
\end{algorithm}
\setlength{\textfloatsep}{6pt}
%
\begin{itemize}
    \item First, we explore possible architectures inside the Cell-A (lines 3-5). 
    Here, the Cell-B architecture is initially set the same as Cell-A (line 6).
    Then, we build an SNN based on Cell-A and Cell-B, and quantize it (lines 7-8).
    \item We evaluate the SNN memory cost, i.e, number of SNN weight parameters (lines 9-10).
    If memory constraint is met, then we perform the fitness evaluation (line 11). 
    If the fitness score is higher than the saved one, we record the SNN architecture as a solution candidate (lines 12-16).
    \item Second, we explore possible architectures inside the Cell-B (lines 3, 20-21). 
    Then, we build an SNN based on the states of Cell-A and Cell-B, and quantize it (lines 22-23).  
    \item We evaluate the candidate for its memory, area, latency, and energy costs (lines 24-28). 
    If all constraints are met, we perform the fitness function evaluation (line 29).
    If the fitness score is higher than the saved one, then we record the SNN architecture as a solution candidate (lines 30-33).
    \item After all steps are finished, we consider the saved SNN architecture as the NAS solution which is ready to train.
\end{itemize}
Furthermore, to enable quick and accurate area, latency, and energy calculations in the NAS, we also propose adjustments considering the precision of NVM device ($bit_D$) and SNN weights ($bit_W$). 
Specifically, we employ $\lceil bit_W/bit_D \rceil$ as an adjustment factor of the corresponding calculations for allocating the crossbar area and calculating the energy consumption, and hence the processing latency.   

\section{Evaluation Methodology}
\label{Sec_Eval}

\begin{figure}[t]
\centering
\includegraphics[width=\linewidth]{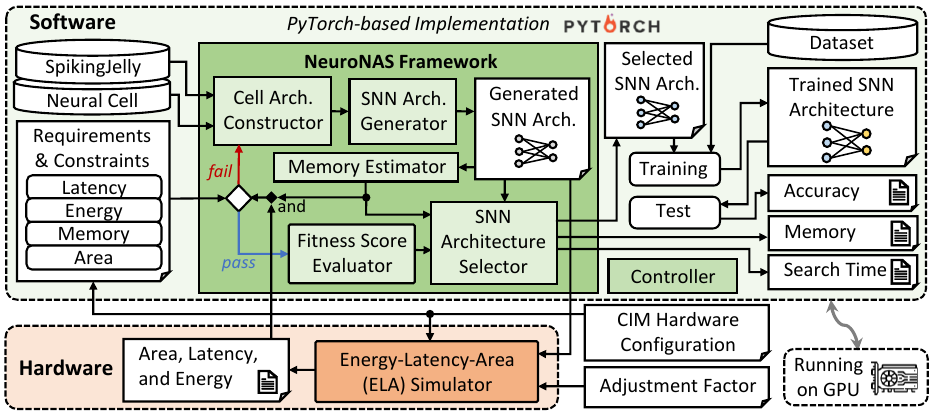}
\vspace{-0.6cm}
\caption{The experimental setup and tools flow.}
\label{Fig_Eval_Method}
\end{figure}

We build an experimental setup and tools flow as shown in Fig.~\ref{Fig_Eval_Method}, while considering the widely-used evaluation settings in the SNN community~\cite{Ref_Nunes_SNNsurvey_Access22, Ref_Putra_FSpiNN_TCAD20, Ref_Li_DynTimestepSNNs_DAC23, Ref_Minhas_SurveyNCL_arXiv24}.
For design constraints, we consider 10M parameters for memory, 1000$mm^2$ for area, 500$ms$ for latency, and 1000$\mu J$ for energy budgets, representing the application requirements and constraints for mobile agent applications.

\textbf{Software Evaluation:}
We realize the NeuroNAS framework using PyTorch implementation based on the SpikingJelly~\cite{Ref_Fang_SpikingJelly_SciAdv23}, then run it on the Nvidia GeForce RTX A6000 Multi-GPU machines. 
We train the network using a surrogate gradient learning with 200 epochs.
To show the generality of our framework, we use CIFAR-10, CIFAR-100, and TinyImageNet-200 which represent simple, medium, and complex datasets, respectively.
For the comparison partner, we consider the state-of-the-art SNASNet with its default settings (e.g., 5000x random search iterations)~\cite{Ref_Kim_SNASNet_ECCV22}. 
We use a 32-bit fixed-point (FxP32) for SNASNet implementation on IMC hardware, since there is no quantization in SNASNet.
The outputs include accuracy, memory, and searching time.

\textbf{Hardware Evaluation:}
To evaluate area, processing latency, and energy consumption, we employ the state-of-the-art energy-latency-area (ELA) simulator while considering the SpikeFlow IMC hardware architecture with real measurements-based RRAM device values from~\cite{Ref_Moitra_SpikeSim_TCAD23}.
The circuit and device parameters are shown in Table~\ref{Tab_CIMvalues}.

\begin{table}[t]
\caption{The IMC circuit and NVM (RRAM) device parameters considered in the ELA simulator; based on studies in~\cite{Ref_Moitra_SpikeSim_TCAD23}.}
\vspace{-0.2cm}
\centering
\scriptsize
\begin{tabular}{|c|c?c|c|c|} 
\hline
\multicolumn{4}{|c|}{\textbf{Circuit Parameters}} \\ 
\hline
\hline
NoC Topology \& Width & Mesh \& 32bits& Size of GBuff & 20KB \\
Clock Frequency & 250MHz & Size of TBuff & 10KB \\
Number of Crossbar-per-PE & 9 & Size of PBuff & 5KB \\
Number of PE-per-Tile & 8 & Size of TIB  & 50KB \\
Multiplexer Size & 8 & Size of PIB & 30KB \\
Precision of Crossbar ADC & 4bits & $V_{DD}$ & 0.9V \\
Column Parasitic Resistance & 5$\Omega$ & $V_{read}$ & 0.1V \\
\hline
\hline
\multicolumn{4}{|c|}{\textbf{RRAM Device Parameters}~\cite{Ref_Hajri_RRAM_Access19} } \\ 
\hline
Bit(s)-per-Cell & 1 & $R_{on}$ & 20k$\Omega$ \\
Write Variation & 0.1 & $R_{off}$ & 200k$\Omega$ \\
\hline
\end{tabular}
\label{Tab_CIMvalues}
\end{table}
\setlength{\textfloatsep}{4pt}

\section{Results and Discussion}
\label{Sec_Results}

\begin{figure*}[t]
\centering
\includegraphics[width=\linewidth]{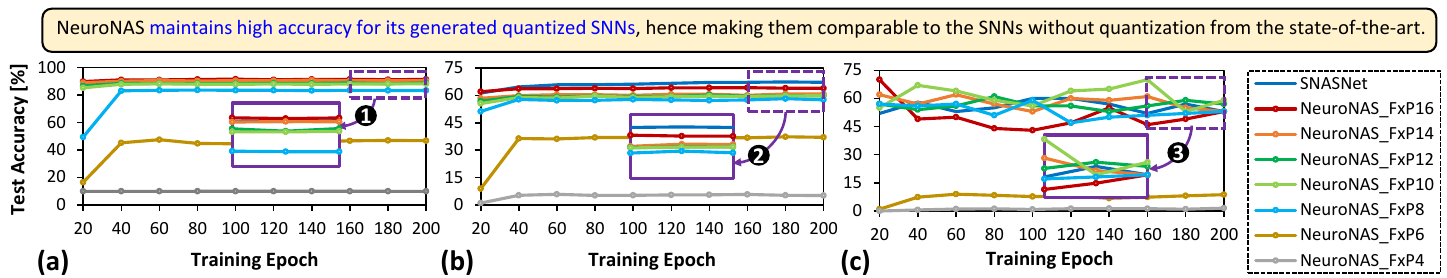}
\vspace{-0.7cm}
\caption{Results of the test accuracy achieved by the SNASNet and our NeuroNAS across training epochs and different precision levels (i.e., FxP16, FxP14, FxP12, FxP10, FxP8, FxP6, and FxP14) and different datasets: \textbf{(a)} CIFAR-10, \textbf{(b)} CIFAR-100, and \textbf{(c)} TinyImageNet-200.}
\label{Fig_Results_Accuracy}
\vspace{-0.1cm}
\end{figure*}

\begin{figure*}[t]
\centering
\includegraphics[width=\linewidth]{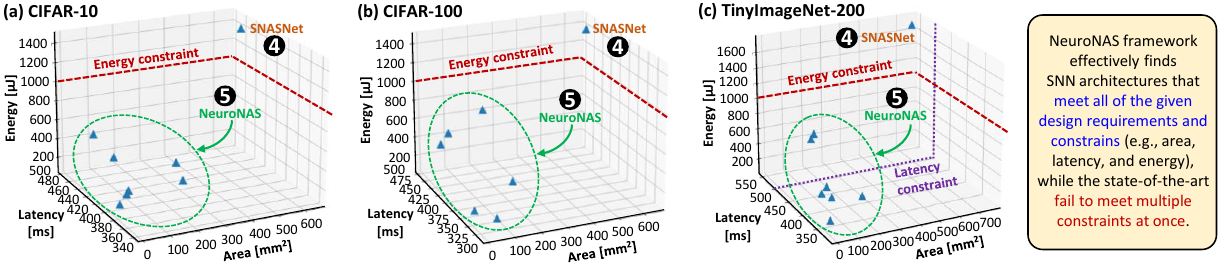}
\vspace{-0.7cm}
\caption{Results of the hardware area, processing latency, and energy consumption achieved by the SNASNet and our NeuroNAS across different precision levels (i.e., FxP16, FxP14, FxP12, FxP10, FxP8, FxP6, and FxP14) and different datasets: \textbf{(a)} CIFAR-10, \textbf{(b)} CIFAR-100, and \textbf{(c)} TinyImageNet-200.}
\label{Fig_Results_3Metrics}
\vspace{-0.5cm}
\end{figure*}

\begin{figure}[t]
\centering
\includegraphics[width=\linewidth]{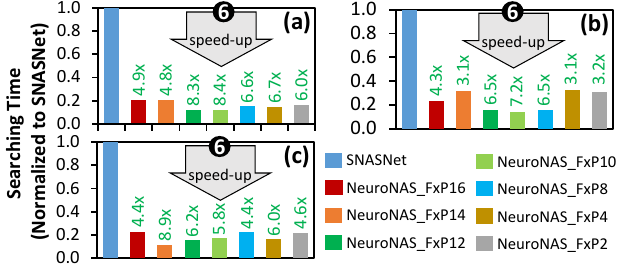}
\vspace{-0.7cm}
\caption{Results of the searching time achieved by the SNASNet and our NeuroNAS across different precision levels (i.e., FxP16, FxP14, FxP12, FxP10, FxP8, FxP6, and FxP14) and different datasets: \textbf{(a)} CIFAR-10, \textbf{(b)} CIFAR-100, and \textbf{(c)} TinyImageNet-200.}
\label{Fig_Results_SearchTime}
\end{figure}

\subsection{Maintaining High Accuracy}
\label{Sec_Results_Accuracy}

Fig.~\ref{Fig_Results_Accuracy} provides the experimental results for accuracy.
Here, the state-of-the-art (i.e., SNASNet) achieves 90.79\%, 67.18\%, and 53\% accuracy for CIFAR-10, CIFAR-100, and TinyImageNet-200, respectively. 
These are obtained through the network generation in SNASNet, which simply relies on the random search iteration. 
Meanwhile, our NeuroNAS can achieve comparable accuracy to the state-of-the-art with smaller precision levels (i.e., with at least 8bits), as shown by \circledB{1} for CIFAR-10, \circledB{2} for CIFAR-100, and \circledB{3} for TinyImageNet-200. 
For CIFAR-10, NeuroNAS achieves accuracy of 91.44\% (FxP16), 90.53\% (FxP14), 88.92\% (FxP12), 88.27\% (FxP10), and 83.44\% (FxP8). 
For CIFAR-100, NeuroNAS achieves accuracy of 63.90\% (FxP16), 60.74\% (FxP14), 59.87\% (FxP12), 59.83\% (FxP10), and 57.59\% (FxP8). 
Meanwhile, for TinyImageNet-200, NeuroNAS achieves accuracy of 53\% (FxP16), 53\% (FxP14), 57\% (FxP12), 59\% (FxP10), and 53\% (FxP8). 
These results show that, our NeuroNAS preserves high accuracy comparable to that of the state-of-the-art across different datasets while employing smaller precision levels. 
Moreover, in the TinyImageNet-200 case, NeuroNAS can consistently achieve similar or even higher accuracy to the state-of-the-art. 
The reasons are the following: (1) NeuroNAS only employs cell operations that have high significance and positive impact on the accuracy, and (2) NeuroNAS constructs the network by considering its fitness score under quantized weights, hence enabling more accurate evaluation of network candidates and search toward the desired SNN architecture. 

\subsection{Ensuring Hardware-aware SNN Generation }
\label{Sec_Results_Constraints}

Fig.~\ref{Fig_Results_3Metrics} provides the experimental results for memory, area, latency, and energy consumption.

\smallskip
\textbf{Memory:} 
Both SNASNet and our NeuroNAS satisfy the memory constraint. 
SNASNet incurs 1.28M weight parameters for CIFAR-10, and 1.93M for CIFAR-100 and TinyImageNet-200. 
Meanwhile, NeuroNAS incurs 888K-4.7M weight parameters for CIFAR-10, 1.77M-4.72M for CIFAR-100, and  1.77M-3.83M for TinyImageNet-200 across different precisions (i.e., FxP4-FxP16).
These are obtained as NeuroNAS incorporates the memory constraint in its NAS.

\smallskip
\textbf{Area:}
SNASNet generates networks that occupy area of 634mm$^2$ for CIFAR-10 and CIFAR-100, and 755mm$^2$ for TinyImageNet-200; see \circledB{4}.
Since SNASNet does not consider the area constraint, the hardware area of the generated network is not under control during its NAS process.
Moreover, in each neural cell, there are multiple convolutional (CONV) layers which cannot be mapped in the same IMC tile due to the mapping strategy. This mapping forces the IMC processing to utilize more resources (e.g., IMC tiles).
Meanwhile, our NeuroNAS generates networks that occupy area of 26mm$^2$-187mm$^2$ for CIFAR-10 (70\%-96\% savings from SNASNet), 25mm$^2$-209mm$^2$ for CIFAR-100 (67\%-96\% savings from SNASNet), and 25mm$^2$-166mm$^2$ for TinyImageNet-200 (78\%-96\% savings from SNASNet) across different precision levels; see \circledB{5}. 
These indicate that \textit{NeuroNAS can ensure the generated network to meet the area constraint}, which also leads to smaller area costs compared to SNASNet. 
The reason is that, NeuroNAS includes the area constraint in its NAS, making it possible to develop both networks and IMC hardware within a custom area budget.

\smallskip
\textbf{Latency:}
SNASNet generates networks that have latency of 468ms for both CIFAR-10 and CIFAR-100, and 545ms for TinyImageNet-200, which does not meet the latency constraint; see \circledB{4}.
The reason is that, SNASNet does not consider the latency budget in its NAS process.
Meanwhile, our NeuroNAS generates networks that have latency of 340ms-494ms for CIFAR-10, 315ms-494mm for CIFAR-100 and TinyImageNet-200 across different precisions; see \circledB{5}.
These show that \textit{NeuroNAS can ensure the generated network to fulfill the latency constraint}, since it considers the latency information in its NAS, making it possible to develop both networks and IMC hardware within a custom latency budget.

\smallskip
\textbf{Energy Consumption:}
SNASNet generates networks that incur energy consumption about 1425$\mu J$ for both CIFAR-10 and CIFAR-100, and 1696$\mu J$ for TinyImageNet-200, hence they do not meet the energy budget; see \circledB{4}.
The reason is that, SNASNet does not consider the energy budget in its NAS process and focuses only on accuracy.
Moreover, in each neural cell, there are multiple CONV layers which cannot be mapped in the same IMC tile due to the mapping strategy, which forces the IMC processing to utilize more compute resources (e.g., IMC tiles), which in turn leading to higher energy consumption.
Our NeuroNAS addresses these limitations by incorporating the energy budget in its NAS, hence avoiding the energy consumption to surpass the given constraint.
Specifically, NeuroNAS generates networks that incur energy consumption of 127$\mu J$-576$\mu J$ for CIFAR-10 (59\%-91\% savings from SNASNet), 110$\mu J$-712$\mu J$ for CIFAR-100 (50\%-92\% savings from SNASNet), and 110$\mu J$-644$\mu J$ for TinyImageNet-200 (74\%-93\% savings from SNASNet) across different precision levels; see \circledB{5}.

\subsection{Searching Time Speed-Up}
\label{Sec_Results_SearchTime}

Fig.~\ref{Fig_Results_SearchTime} presents the experimental results of the searching time.
In general, our NeuroNAS offers significant speed-ups for the searching time as compared to the state-of-the-art. 
Specifically, NeuroNAS improves the searching time by 4.8x-8.4x for CIFAR-10, by 3.1x-7.2x for CIFAR-100, and by 4.4x-8.9x for TinyImageNet-200; see \circledB{6}.
The reason is that, our NeuroNAS employs a fewer SNN operations, minimizes redundant architecture candidates to evaluate, incorporates multiple constraints into the search process, hence leading to a smaller search space and a faster searching time as compared to the state-of-the-art.

\subsection{Comparison to Conventional ANN-based Solutions}
\label{Sec_Results_CompareANNs}

In ANN domain, the prominent state-of-the-art work in ``\textit{NAS without training}'' that considers the same benchmark with neural cell-based strategy (i.e., NAS-Bench-201~\cite{Ref_Dong_NASbench201_ICLR20}) is NASWOT~\cite{Ref_Mellor_NASnoTraining_ICML21}. 
We observe that NASWOT with a 32-bit precision can achieve around 92.81\% accuracy for CIFAR-10 and around 69.35\% accuracy for CIFAR-100. 
Meanwhile, NeuroNAS with a 32-bit precision achieves around 91.58\% for CIFAR-10 and around 64.04\% accuracy for CIFAR-100. 
These results indicate that, our NeuroNAS can obtain competitive accuracy as compared to the conventional ANNs because of its effective searching strategy, thus providing an SNN architecture that can learn well during training process.

\begin{figure}[t]
\centering
\includegraphics[width=\linewidth]{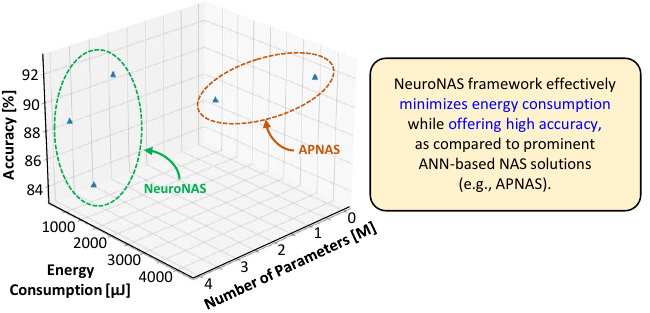}
\vspace{-0.6cm}
\caption{Trade-offs among accuracy, network size (number of parameters), and energy consumption for ANN-based solutions and our NeuroNAS.}
\label{Fig_Results_vsANN}
\end{figure}

Besides competitive accuracy, NeuroNAS also offers significantly higher energy efficiency than ANN-based solutions due to the nature of event-based SNN processing as well as the reduced off-chip data movements, which are maximized through the employment of IMC hardware; see Table~\ref{Tab_Compare} and Fig.~\ref{Fig_Results_vsANN}.
Specifically, our NeuroNAS\_FxP16 can achieve 86\% energy saving while maintaining accuracy within 2\% from APNAS96~\cite{Ref_Achararit_APNAS_Access20} for CIFAR-10 dataset.    

\begin{table}[h]
\caption{Comparison to ANN-based solutions from APNAS~\cite{Ref_Achararit_APNAS_Access20} for network size, accuracy, and energy consumption considering the CIFAR-10 dataset and 65nm technology nodes.}
\centering
\scriptsize
\begin{tabular}{|c|c|c|c|c|}
\hline
\textbf{Domain} & \begin{tabular}[c]{@{}c@{}}\textbf{NAS Method}\end{tabular} & \begin{tabular}[c]{@{}c@{}}\textbf{Number of }\\ \textbf{Parameters}\end{tabular} & \textbf{Accuracy} & \begin{tabular}[c]{@{}c@{}}\textbf{Energy}\\ \textbf{Consumption}\end{tabular} \\ \hline \hline
\multirow{2}{*}{ANN} & APNAS96~\cite{Ref_Achararit_APNAS_Access20} & 0.7M & 92.67\% & 4220.34$\mu J$ \\ 
 & APNAS36~\cite{Ref_Achararit_APNAS_Access20} & 0.1M & 87.02\% & 603.09$\mu J$ \\ \hline
\multirow{3}{*}{SNN} & NeuroNAS\_FxP16 & 2.9M & 91.44\% & 576.89$\mu J$ \\ 
 & NeuroNAS\_FxP12 & 3.9M & 88.92\% & 381.77$\mu J$ \\ 
 & NeuroNAS\_FxP8 & 3.2M & 83.44\% & 220.58$\mu J$ \\ \hline
\end{tabular}
\label{Tab_Compare}
\end{table}
\setlength{\textfloatsep}{4pt}

\section{Conclusion}
\label{Sec_Conclusion}

We propose the novel NeuroNAS framework to develop an SNN architecture for a neuromorphic IMC platfom that can achieve high accuracy under multiple requirements and constraints. 
Its key steps include: (1) optimizing SNN operations, (2) developing an efficient SNN using novel QaFE policy, and (3) employing a hardware-aware search algorithm that incorporates all constraints in its NAS process. 
As a result, NeuroNAS maintains high accuracy comparable to the state-of-the-art while improving other aspects to fulfill all given constraints. 
For instance, NeuroNAS\_FxP8 achieves up to 92\% area savings, 1.2x latency improvements, 84\% energy savings, and 6.6x search time speed-ups across all datasets.
Therefore, NeuroNAS framework enables an effective design automation for streamlining the development of energy-efficient neuromorphic IMC for intelligent mobile agents. 


% ...

\bibliographystyle{IEEEtran}
\bibliography{bibliography.bib}

\begin{thebibliography}{10}
\providecommand{\url}[1]{#1}
\csname url@samestyle\endcsname
\providecommand{\newblock}{\relax}
\providecommand{\bibinfo}[2]{#2}
\providecommand{\BIBentrySTDinterwordspacing}{\spaceskip=0pt\relax}
\providecommand{\BIBentryALTinterwordstretchfactor}{4}
\providecommand{\BIBentryALTinterwordspacing}{\spaceskip=\fontdimen2\font plus
\BIBentryALTinterwordstretchfactor\fontdimen3\font minus \fontdimen4\font\relax}
\providecommand{\BIBforeignlanguage}[2]{{%
\expandafter\ifx\csname l@#1\endcsname\relax
\typeout{** WARNING: IEEEtran.bst: No hyphenation pattern has been}%
\typeout{** loaded for the language `#1'. Using the pattern for}%
\typeout{** the default language instead.}%
\else
\language=\csname l@#1\endcsname
\fi
#2}}
\providecommand{\BIBdecl}{\relax}
\BIBdecl

\bibitem{Ref_Putra_EmbodiedNAI_ICARCV24}
R.~V.~W. Putra, A.~Marchisio, F.~Zayer, J.~Dias, and M.~Shafique, ``Embodied neuromorphic artificial intelligence for robotics: Perspectives, challenges, and research development stack,'' in \emph{2024 18th International Conference on Control, Automation, Robotics and Vision (ICARCV)}, 2024, pp. 612--619.

\bibitem{Ref_Basu_SNNicSurvey_CICC22}
A.~Basu, L.~Deng, C.~Frenkel, and X.~Zhang, ``Spiking neural network integrated circuits: A review of trends and future directions,'' in \emph{2022 IEEE Custom Integrated Circuits Conference (CICC)}, 2022, pp. 1--8.

\bibitem{Ref_Ankit_Resparc_DAC17}
A.~Ankit, A.~Sengupta, P.~Panda, and K.~Roy, ``Resparc: A reconfigurable and energy-efficient architecture with memristive crossbars for deep spiking neural networks,'' in \emph{IEEE/ACM Design Automation Conference (DAC)}, 2017, pp. 1--6.

\bibitem{Ref_Moitra_SpikeSim_TCAD23}
A.~Moitra, A.~Bhattacharjee, R.~Kuang, G.~Krishnan, Y.~Cao, and P.~Panda, ``Spikesim: An end-to-end compute-in-memory hardware evaluation tool for benchmarking spiking neural networks,'' \emph{IEEE Transactions on Computer-Aided Design of Integrated Circuits and Systems (TCAD)}, vol.~42, no.~11, pp. 3815--3828, 2023.

\bibitem{Ref_Roy_PEASE_ISLPED17}
A.~Roy, S.~Venkataramani, N.~Gala, S.~Sen, K.~Veezhinathan, and A.~Raghunathan, ``A programmable event-driven architecture for evaluating spiking neural networks,'' in \emph{IEEE/ACM International Symposium on Low Power Electronics and Design (ISLPED)}, July 2017, pp. 1--6.

\bibitem{Ref_Akopyan_TrueNorth_TCAD15}
F.~Akopyan, J.~Sawada, A.~Cassidy, R.~Alvarez-Icaza, J.~Arthur, P.~Merolla, N.~Imam, Y.~Nakamura, P.~Datta, G.-J. Nam, B.~Taba, M.~Beakes, B.~Brezzo, J.~B. Kuang, R.~Manohar, W.~P. Risk, B.~Jackson, and D.~S. Modha, ``Truenorth: Design and tool flow of a 65 mw 1 million neuron programmable neurosynaptic chip,'' \emph{IEEE Transactions on Computer-Aided Design of Integrated Circuits and Systems (TCAD)}, vol.~34, no.~10, pp. 1537--1557, Oct 2015.

\bibitem{Ref_Sen_ApproxSNN_DATE17}
S.~Sen, S.~Venkataramani, and A.~Raghunathan, ``Approximate computing for spiking neural networks,'' in \emph{Design, Automation Test in Europe Conf. Exhibition (DATE)}, March 2017, pp. 193--198.

\bibitem{Ref_Krithivasan_SpikeBundle_ISLPED19}
S.~Krithivasan, S.~Sen, S.~Venkataramani, and A.~Raghunathan, ``Dynamic spike bundling for energy-efficient spiking neural networks,'' in \emph{2019 IEEE/ACM International Symposium on Low Power Electronics and Design (ISLPED)}, July 2019, pp. 1--6.

\bibitem{Ref_Wu_DirectTrainSNNs_AAAI19}
Y.~Wu, L.~Deng, G.~Li, J.~Zhu, Y.~Xie, and L.~Shi, ``Direct training for spiking neural networks: Faster, larger, better,'' in \emph{AAAI Conference on Artificial Intelligence (AAAI)}, vol.~33, no.~01, 2019, pp. 1311--1318.

\bibitem{Ref_Fang_MemTConstantSNNs_ICCV21}
W.~Fang, Z.~Yu, Y.~Chen, T.~Masquelier, T.~Huang, and Y.~Tian, ``Incorporating learnable membrane time constant to enhance learning of spiking neural networks,'' in \emph{IEEE/CVF International Conference on Computer Vision (ICCV)}, 2021, pp. 2661--2671.

\bibitem{Ref_Lee_SpikeBackprop_FNINS20}
C.~Lee, S.~S. Sarwar, P.~Panda, G.~Srinivasan, and K.~Roy, ``Enabling spike-based backpropagation for training deep neural network architectures,'' \emph{Frontiers in Neuroscience (FNINS)}, p. 119, 2020.

\bibitem{Ref_Zheng_LargerSNNs_AAAI21}
H.~Zheng, Y.~Wu, L.~Deng, Y.~Hu, and G.~Li, ``Going deeper with directly-trained larger spiking neural networks,'' in \emph{AAAI Conference on Artificial Intelligence (AAAI)}, vol.~35, no.~12, 2021, pp. 11\,062--11\,070.

\bibitem{Ref_Na_AutoSNN_ICML22}
B.~Na, J.~Mok, S.~Park, D.~Lee, H.~Choe, and S.~Yoon, ``Autosnn: Towards energy-efficient spiking neural networks,'' in \emph{International Conference on Machine Learning (ICML)}, 2022, pp. 16\,253--16\,269.

\bibitem{Ref_Putra_SpikeNAS_arXiv24}
R.~V.~W. Putra and M.~Shafique, ``Spikenas: A fast memory-aware neural architecture search framework for spiking neural network systems,'' \emph{arXiv preprint arXiv:2402.11322}, 2024.

\bibitem{Ref_Kim_SNASNet_ECCV22}
Y.~Kim, Y.~Li, H.~Park, Y.~Venkatesha, and P.~Panda, ``Neural architecture search for spiking neural networks,'' in \emph{European Conference on Computer Vision (ECCV)}, 2022, pp. 36--56.

\bibitem{Ref_Liu_litESNN_arXiv24}
Q.~Liu, J.~Yan, M.~Zhang, G.~Pan, and H.~Li, ``Lite-snn: Designing lightweight and efficient spiking neural network through spatial-temporal compressive network search and joint optimization,'' \emph{arXiv preprint arXiv:2401.14652}, 2024.

\bibitem{Ref_Yan_NAS4SNN_NeuNet24}
J.~Yan, Q.~Liu, M.~Zhang, L.~Feng, D.~Ma, H.~Li, and G.~Pan, ``Efficient spiking neural network design via neural architecture search,'' \emph{Neural Networks}, p. 106172, 2024.

\bibitem{Ref_Pan_EvoNAS4SNN_arxiv23}
W.~Pan, F.~Zhao, G.~Shen, B.~Han, and Y.~Zeng, ``Multi-scale evolutionary neural architecture search for deep spiking neural networks,'' \emph{arXiv preprint:2304.10749}, 2023.

\bibitem{Ref_Che_AutoSpikeformer_arXiv23}
K.~Che, Z.~Zhou, Z.~Ma, W.~Fang, Y.~Chen, S.~Shen, L.~Yuan, and Y.~Tian, ``Auto-spikformer: Spikformer architecture search,'' \emph{arXiv preprint arXiv:2306.00807}, 2023.

\bibitem{Ref_Wang_AutoST_ICASSP24}
Z.~Wang, Q.~Zhao, J.~Cui, X.~Liu, and D.~Xu, ``Autost: Training-free neural architecture search for spiking transformers,'' in \emph{IEEE International Conference on Acoustics, Speech and Signal Processing (ICASSP)}, 2024, pp. 3455--3459.

\bibitem{Ref_Putra_SparkXD_DAC21}
R.~V.~W. Putra, M.~A. Hanif, and M.~Shafique, ``Sparkxd: A framework for resilient and energy-efficient spiking neural network inference using approximate dram,'' in \emph{2021 58th ACM/IEEE Design Automation Conference (DAC)}, 2021, pp. 379--384.

\bibitem{Ref_Putra_SpikeDyn_DAC21}
R.~V.~W. Putra and M.~Shafique, ``Spikedyn: A framework for energy-efficient spiking neural networks with continual and unsupervised learning capabilities in dynamic environments,'' in \emph{2021 58th ACM/IEEE Design Automation Conference (DAC)}, 2021, pp. 1057--1062.

\bibitem{Ref_Putra_ReSpawn_ICCAD21}
R.~V.~W. Putra, M.~A. Hanif, and M.~Shafique, ``Respawn: Energy-efficient fault-tolerance for spiking neural networks considering unreliable memories,'' in \emph{2021 IEEE/ACM International Conference On Computer Aided Design (ICCAD)}, 2021, pp. 1--9.

\bibitem{Ref_Mellor_NASnoTraining_ICML21}
J.~Mellor, J.~Turner, A.~Storkey, and E.~J. Crowley, ``Neural architecture search without training,'' in \emph{International Conference on Machine Learning (ICML)}, 2021, pp. 7588--7598.

\bibitem{Ref_Dong_NASbench201_ICLR20}
X.~Dong and Y.~Yang, ``Nas-bench-201: Extending the scope of reproducible neural architecture search,'' in \emph{International Conference on Learning Representations (ICLR)}, 2020.

\bibitem{Ref_Nunes_SNNsurvey_Access22}
J.~D. Nunes, M.~Carvalho, D.~Carneiro, and J.~S. Cardoso, ``Spiking neural networks: A survey,'' \emph{IEEE Access}, vol.~10, 2022.

\bibitem{Ref_Putra_FSpiNN_TCAD20}
R.~V.~W. Putra and M.~Shafique, ``Fspinn: An optimization framework for memory-efficient and energy-efficient spiking neural networks,'' \emph{IEEE Transactions on Computer-Aided Design of Integrated Circuits and Systems (TCAD)}, vol.~39, no.~11, pp. 3601--3613, 2020.

\bibitem{Ref_Li_DynTimestepSNNs_DAC23}
Y.~Li, A.~Moitra, T.~Geller, and P.~Panda, ``Input-aware dynamic timestep spiking neural networks for efficient in-memory computing,'' in \emph{2023 60th ACM/IEEE Design Automation Conference (DAC)}, 2023, pp. 1--6.

\bibitem{Ref_Minhas_SurveyNCL_arXiv24}
M.~F. Minhas, R.~V.~W. Putra, F.~Awwad, O.~Hasan, and M.~Shafique, ``Continual learning with neuromorphic computing: Theories, methods, and applications,'' \emph{arXiv preprint arXiv:2410.09218}, 2024.

\bibitem{Ref_Fang_SpikingJelly_SciAdv23}
W.~Fang, Y.~Chen, J.~Ding, Z.~Yu, T.~Masquelier, D.~Chen, L.~Huang, H.~Zhou, G.~Li, and Y.~Tian, ``Spikingjelly: An open-source machine learning infrastructure platform for spike-based intelligence,'' \emph{Science Advances}, vol.~9, no.~40, 2023.

\bibitem{Ref_Hajri_RRAM_Access19}
B.~Hajri, H.~Aziza, M.~M. Mansour, and A.~Chehab, ``Rram device models: A comparative analysis with experimental validation,'' \emph{IEEE Access}, vol.~7, pp. 168\,963--168\,980, 2019.

\bibitem{Ref_Achararit_APNAS_Access20}
P.~Achararit, M.~A. Hanif, R.~V.~W. Putra, M.~Shafique, and Y.~Hara-Azumi, ``Apnas: Accuracy-and-performance-aware neural architecture search for neural hardware accelerators,'' \emph{IEEE Access}, vol.~8, pp. 165\,319--165\,334, 2020.

\end{thebibliography}

\end{document}